# Co-Evolution of Multi-Robot Controllers and Task Cues for Off-World Open Pit Mining

Jekan Thangavelautham[1] and Yinan Xu[2]

[1]Space and Terrestrial Robotic Exploration (SpaceTREx) Laboratory, University of Arizona, 1130 N Mountain Ave., N635, Tucson, Arizona United States, E-mail: jekan@arizona.edu
[2]Space and Terrestrial Robotic Exploration (SpaceTREx) Laboratory, University of Arizona, 1130 N Mountain Ave., N417, Tucson, Arizona United States, E-mail: yinanx@email.arizona.edu

**ABSTRACT**

Robots are ideal for open-pit mining on the Moon as it is a dull, dirty, and dangerous task. The challenge is to scale up productivity with an ever-increasing number of robots. This paper presents a novel method for developing scalable controllers for use in multi-robot excavation and site-preparation scenarios. The controller starts with a blank slate and does not require human-authored operations scripts nor detailed modeling of the kinematics and dynamics of the excavator. The 'Artificial Neural Tissue' (ANT) architecture is used as a control system for autonomous robot teams to perform resource gathering. This control architecture combines a variable-topology neural-network structure with a coarse-coding strategy that permits specialized areas to develop in the tissue.

Our work in this field shows that fleets of autonomous decentralized robots have an optimal operating density. Too few robots result in insufficient labor, while too many robots cause *antagonism*, where the robots undo each other's work and are stuck in gridlock. In this paper, we explore the use of templates and task cues to improve group performance further and minimize antagonism. Our results show light beacons and task cues are effective in sparking new and innovative solutions at improving robot performance when placed under stressful situations such as severe time-constraint.

## 1 INTRODUCTION

The use of multiple robots to autonomously perform labor-intensive tasks in dangerous or inaccessible environments has many useful applications in search-rescue, construction, site-preparation/clearing, resource exploration, and mining. The use of multi-robot systems to design, develop, and operate a mining base on the surface of the Moon could be a game-changer in kickstarting a new space economy. The resources mined could include water and metals [1]. The surface of the Moon faces extremes in temperature, cosmic and solar radiation, and periodic micro-meteorite impacts that pose high risks for human workers. In addition, the lunar regolith is extremely abrasive, and the vacuum conditions all make conditions further inhospitable. Lunar base structures could be built to shelter human workers on the base but require substantial additional resources compared to the use of robot teams.

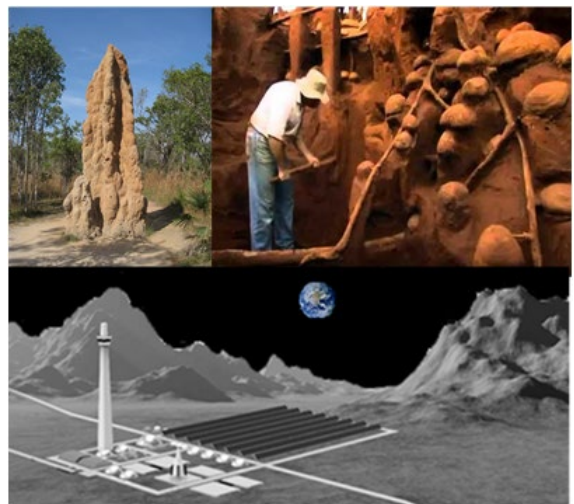

*Figure 1: The organization principles of cathedral termite mounds (top left) and ant cities (top right) could be mimicked to construct and operate future robot mining bases on the Moon.*

The decentralized control of robot teams has many potential advantages, including fault tolerance, parallelism, reliability, scalability, and simplicity in agent design [2]. Challenges remain in scaling-up these decentralized robot teams to ever larger and complex tasks with increased parallelism. A wrong choice of coordination and communication behaviors limits overall system performance and result in *antagonism* when multiple individuals trying to perform the same task interfere with one another, causing slow and unreliable group performance or, worse, gridlock. It is marvelous to observe social insects such as ants, termites, and bees, without any centralized coordination or control, produce remarkably robust, scalable group behaviors to build sophisticated structures including tunnels, hives, and cathedral mounds containing internal heating and cooling systems [3]. A self-sufficient termite mound that can harvest its own food in the arid Australian Desert is an inspiration for a fully autonomous robotic mining base on the Moon

Conventional model and behavior-based control require domain knowledge of a task at hand. It is often unclear how best to organize and control the individuals to perform the required group behaviors. Work by [4] relies on task-specific human knowledge to develop simple ``if-then'' rules and coordination behaviors to solve multi-robot tasks. The required coordination and communication behaviors are often unintuitive and not well understood. Domain knowledge and a priori experience are often relied upon up to determine whether its best to assemble the robots into a formation or distribute them throughout the work area. Decentralized methods use an automated algorithm to perform the partitioning but require imposing physical boundaries or requiring mapping and localization capability in an unstructured setting. It's typically unclear if robots need to be in formation or distributed throughout the work area for the entirety of a task. Typically, a human supervisor determines the coordination behaviors manually through trial and error, which can be tedious and lacks direction.

In the absence of domain knowledge, human designers learn the required multi-robot coordination strategies from a process of trial and error. Our approach takes this to the next incremental step using an artificial Darwinian approach to automate the controller design process. In this work, the problem is addressed using an artificially evolvable neural network architecture that draws heavily from nature called the Artificial Neural Tissue framework [5-6]. The Artificial Neural Tissue (ANT) is an adaptive approach that learns to solves tasks through trial and error. ANT superimposes on a typical feed-forward or recurrent neural network, a novel mapping scheme that can selectively activate and inhibit the genome (genotype) and the neural network (phenotype). Selective activation and inhibition of the phenotype are performed dynamically using a coarse-coding [7-8] mechanism, resulting in modules of neurons to be added, removed, and rewired. During training, these modules specialize and handle separate segments of a task, thus performing task decomposition. The coarse coding mechanism has a biological analogy as neurons can communicate not only electrically by exchanging signals along wires (axons) but also wirelessly through chemical diffusion [9].

In this paper, we provide expanded analysis of the resource collection task from our previous work [10]. The effort is motivated by plans for terrestrial open pit mining and site preparation on the lunar surface (Figure 1). Here we show the approach can produce scale-able controllers that require only a global fitness (goal) function, a generic set of behavior primitives, sensory input, and a training environment. In this paper, we evaluate the role of highly-organized cooperative behavior and environmental cues in advancing the scalability of multi-robot systems. In an earlier work [10], we observed the emergence of highly organized cooperative behaviors include bucket-brigades, where robots pass material from one to another until the material has reached its end goal. Bypassing material from robot to robot, each robot minimizes travel distance, and the burden is shared among the collective. Here we perform experiments to determine conditions under which bucket-brigade behaviors emerge. Next, we compare these highly organized cooperative strategies to the use of environmental cues that could be used to facilitate partitioning of a complex task to multiple agents operating independently and in parallel and analyze overall effectiveness in robot scalability. In section 2, we present the background to the problems at hand, followed by a description of the Artificial Neural Tissue (ANT) framework in Section 3, a description of the robotic tasks in Section 4, followed by results and discussion in Section 5 and conclusions in Section 6.

## 2 BACKGROUND

Insect societies, including termites, ants, and bees, provide an excellent model for designing a decentralized multi-robot system. Some of the identified mechanisms within these insect societies are used to perform cooperative behaviors. They include the use of templates, stigmergy, and self-organization. *Templates* are environmental cues that trigger individual behaviors [11]. *Stigmergy* is a form of implicit communication that is mediated through the environment [12]. *Self-organization* describes how local or individual behaviors give rise to a global structure in a system that is not in equilibrium [13]. Roboticists have attempted to adapt these mechanisms for use in real-world scenarios with mixed success.

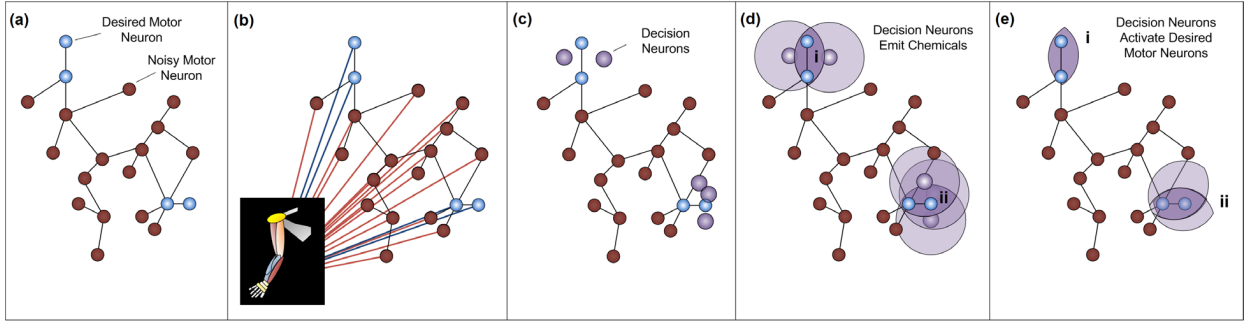

*Figure 2: In a randomly generated tissue, most motor neurons would produce spurious/incoherent output (a) that would `drown out' signals from a few desired motor neurons due to spatial crosstalk [18] (b). This can make training intractable for difficult tasks. Neurotransmitter (chemicals) emitted by decision neurons (c) selectively activate networks of desired motor neurons in shaded regions (i) and (ii) by coarse-coding overlapping diffusion fields as shown (d). This inhibits noisy motor neurons and eliminates spatial crosstalk (e).*

However, many existing approaches suffer from another emergent feature called *antagonism* when multiple agents in the collective undo each other's work reducing overall performance or worse, resulting in gridlock. Our approach is evolutionary in nature, and it is able to 'learn' how to take advantage of the techniques identified above and improve system-level performance.

Our system learns to mitigate the effects of *antagonism*, which is difficult to do in hand-crafted systems. In insect societies, templates may be a natural event or generated by the collective themselves. They may include gradients in light, humidity, temperature, and chemical substances. In robotics, template-based methods include the use of light fields to direct the creation of circular [14] and linear walls and planar annulus structures [15]. Spatiotemporally varying templates allow the creation of more complex structures for construction [16].

Stigmergy describes physical changes in the environment that are used to mediate communication. The use of pheromone triggered events is a common example. Stigmergy has been widely attempted in collective-robotic, including for construction tasks, such as blind bulldozing [20], box pushing [4], heap formation [19], and tiling pattern formation [21]. These examples model the mechanism with user-defined deterministic 'if-then' rules or stochastic behaviors. Often the goal is to develop a minimalist controller for agents that have access only to local information but can work together to achieve a common objective. This is challenging to do manually since the global effect of local interactions is often hard to predict. Early attempts started with the design of a single controller, which minimizes interaction with other robots, thus treating them as 'obstacles' to be avoided. Using multiple robots, the workspace is partitioned for each robot, thus enabling scalability.

Often robots are designed with a single individual in mind and multiplied. When one robot needs to interact with another, a set of arbitration rules is developed with limited domain knowledge. The result is that the interactions between agents are more often antagonistic than cooperative. Some of the latest techniques attempt to achieve a desired global behavior using potential functions [17]. These techniques require more supervision and impose increased requirements on the robots, particularly accurate localization systems. The challenge with potential functions is that they are not intuitive and require prior experience and the ability to perform trial and error testing. Designing these controllers by hand can devolve into a process of trial and error, but without a systematic approach to measure net progress. Our research has focused on automating stochastic trial and error searches by measuring progress using a fitness function that measures system performance.

## 3    ARTIFICIAL NEURAL TISSUE

The ANT framework presented here is a bio-inspired approach that simultaneously addresses both the problems in designing rule-based systems by hand ANT uses a coarse coding mechanism to dynamically activate and inhibit networks of neurons. This reduces the number of active, interconnected neurons, thus reducing spatial crosstalk. This, as will be presented later in the paper, provides good scalability and generalization of sensory input [5-6]. ANT starts with a blank slate. It evolves controllers to optimize overall system performance using a goal function. As we show in this work, the Darwinian training process, if given the right opportunity, can exploit templates, stigmergy, self-organization, and even mitigate the effects of antagonism.

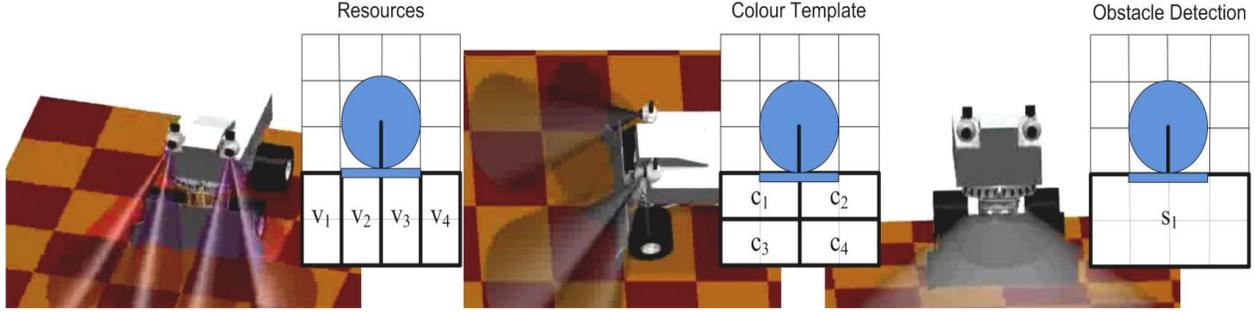

*Figure 4: Input sensor mapping, with simulation model inset.*

ANT starts with a developmental program encoded in an artificial 'genome' that constructs a three-dimensional neural tissue and regulatory functions. The tissue consists of two types of units, *decision neurons,* and *motor-control neurons*. Let's begin with a tissue consisting of motor control neurons connected electrically by wires (Figure 2a). By probability, most of these neurons will produce incoherent output, while a few may produce the desired feature. If the signals from all of these neurons are integrated, then these 'noisy' neurons would drown out the desired output signal (Figure 2b) due to spatial crosstalk [18]. Within ANT, decision neurons emit chemicals that diffuse omnidirectionally (shown shaded) (Figure 2d). By coarse-coding multiple overlapping diffusion fields, the desired motor neuron is selected, and neighboring 'noisy' neurons are inhibited, referred to as neural regulation. With multiple overlapping diffusion fields (Figure 2d ii), there is inherent redundancy in the tissue, and when one decision neuron is damaged, the desired motor neurons could still be selected. Readers are referred to [5-6], where there is a complete description of ANT.

## 4  RESOURCE COLLECTION TASK

The utility of the ANT controller is demonstrated in simulation on the resource-collection task. Modeled after open-pit mining scenarios, a team of robots collects resource material distributed throughout a work area and deposits it in a designated dumping area (Figure 3). The work area is modeled as a 2D grid with one robot occupying four grid squares. Each robot controller must attain a number of functions, including avoiding collision with other robots,

A credible solution possesses a number of capabilities, including gathering resource material, avoiding work area perimeter, avoiding collisions with other robots, and piling resources into a mound at the designated dumping location. The dumping area has perimeter markings on the floor, and a navigational light beacon mounted nearby. The two colors on the border are intended to allow the controller to distinguish whether it is inside or outside the dumping area. The robot may choose to use the light beacon to track the dumping area which is efficient. Without it, the robots need to randomly search for the blue perimeter and search along it to find the dumping area which is more time consuming. The fitness function measures the amount of resource material collected in the dumping area after a finite number of time steps averaged over 100 different initial conditions.

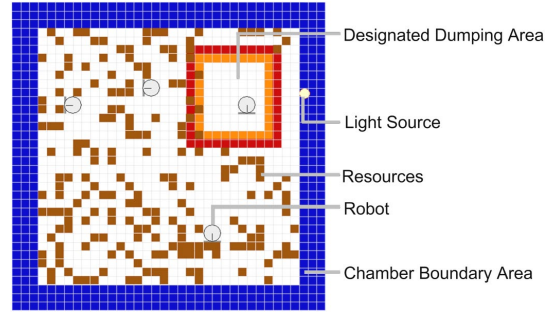

*Figure 3: 2D grid world model of open-pit mining workspace.*

Table 1: Robot Sensor Inputs

| Sensor Variable | Function | Description |
|---|---|---|
| $V_1…V_4$ | Resource Detection | Resource, No Resource |
| $C_1…C_4$ | Template Detection | Blue, Red, Orange, Floor |
| $S_1, S_2$ | Obstacle Detection | Obstacle, No Obstacle |
| $LP_1$ | Light Position | Left, Right, Center |
| $LD_1$ | Light Range | 0-10 (distance to light) |

Table 2: Robot Basis Behavior

| Order | Behavior | Description |
|---|---|---|
| 1 | Dump Resource | Move one grid square back; turn left |
| 2 | Move Forward | Move one grid square forward |

| 3 | Turn Right | Turn 90º right |
| 4 | Turn Left | Turn 90º left |
| 5,7,9,11 | Bit Set | Set memory bit i to 1, i=1...4 |
| 6,8,10,12 | Bit Clear | Set memory bit i to 0, i=1...4 |

Inputs to the ANT controller are shown in Table 1. The sensor input is used to detect resources, templates, obstacles, and a light beacon. The robots are modeled on a fleet of Argo rovers designed and built at the University of Toronto Institute for Aerospace Studies. A layout of the sensor inputs is shown in Figure 4. The robots are equipped with a pair of webcams and sonars. All raw input data are discretized. The sonar sensors are used to determine the values of $S_1$ and $S_2$. One of the cameras is used to detect resource material and colored floor templates. The other camera is used to track the light beacon. To identify resources and colored floor templates, a Naive Bayes classifier is used to perform color recognition [9]. Basic feature-detection heuristics are used to determine the values of $V_1…V_4$ and $C_1…C_4$ based on the grid locations shown. The light beacon tracked by adjusting the camera shutter and gain to ensure that the light source is visible while other work area features are underexposed. The relative position of the light source, $LP_1$, is determined based on camera pan angle. The distance to the light source $LD_1$ is calculated based on its size in the tracking image. The robots can read four memory bits and that can be manipulated using some of the basis behaviors or behavior primitives. Together there are $2^4 \times 4^4 \times 2^2 \times 3 \times 11 \times 2^4 = 8.7 \times 10^6$ possible sensor inputs.

Table 2 lists a pre-ordered set of basis behaviors available for the robot to execute. These behaviors are activated depending on the output of ANT, and all occur within a single timestep. Noting that each behavior in Table 2 can be triggered or not for any one of $8.7 \times 10^6$ possible combinations of sensor inputs, there is a total of $2^{12 \times 8.7 \times 10^6} \approx 10^{3 \times 10^7}$ possible states in the search space! Alternatively, in some experiments, ANT uses a set of behavior primitives that involves turning on a combination of left and right motor to produce the basis behaviors in Table 2, in addition to eight more behaviors, which include slide left or right one square, move diagonally left or right one square and two instances of pivot left or right. In addition, the order of behavior execution is an evolved parameter.

Task decomposition is typically needed to tackle very large search spaces and find desired solutions. ANT using its coarse-coding scheme described earlier is shown to perform task decomposition [6] and tackle this resource collection task with its large task space. ANT controllers are first evolved in a simplified training environment. The evolutionary algorithm population size for training is P=100, with crossover probability $p_c$=0.7, mutation probability $p_m$=0.025, and a tournament size of 0.06P. The tissue is initialized as a 'seed culture,' with $3 \times 6$ motor control neurons in one layer and pre-grown to include between 10-110 neurons (selected randomly) before starting training. Initializing with a random number of neurons was found to produce a diverse initial population irrespective of the task being trained for.

## 5 RESULTS AND DISCUSSION

Figure 5 shows the fitness (population best) of ANT controllers evaluated at each generation of the artificial evolutionary process. Figure 6 shows an evolved solution. The results show that system performance increases with the number of robots. With more robots, each robot has a smaller area to cover in trying to gather and dump resources.

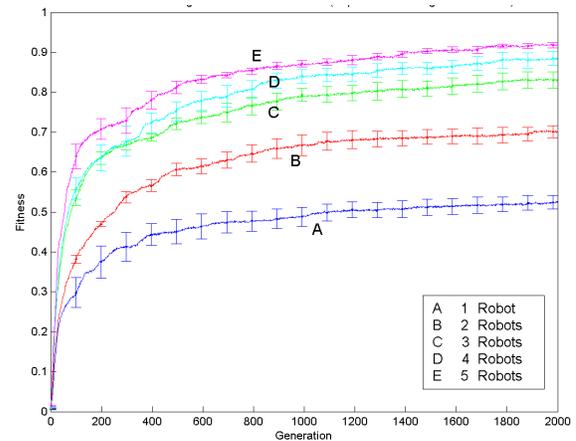

*Figure 5: Typical evolutionary run using the ANT controllers. Increased number of robots see improvement in fitness up to an optimal density.*

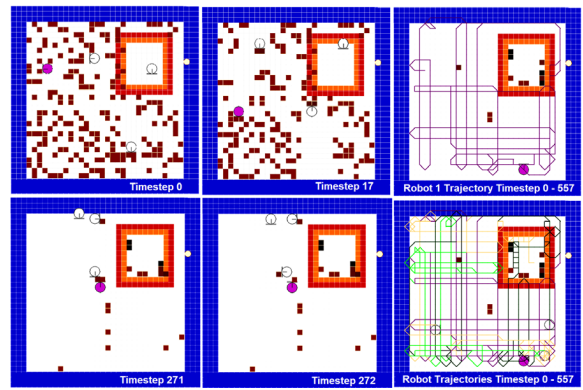

*Figure 6: Simulation of a team four robots completing the resource collection task. Snapshots are taken at 0, 17, 271, and 272 timesteps; robot trajectories*

*are shown top and bottom right. At timestep 271 shows evidence for bucket-brigade behavior.*

The simulation runs indicate that a point of diminishing returns is eventually reached. Beyond this point, additional robots have a minimal or adverse effect on system performance, with the initial resource density and robot density kept constant. The evolutionary process enables the decomposition of a goal task based on global fitness, and the tuning of behaviors depending on the robot density. Snapshots of an ANT controller solution completing the resource collection task are shown in Figure 6.

### 5.1 Evolvable Controller Comparison

Having seen that ANT can solve the resource collection task, we compare ANT to other methods, including standard fixed topology neural networks and variable topology NEAT [22] in Figure 7. For this comparison, topologies for NEAT and fixed neural networks are randomly generated and contain between 10 and 110 neurons to make the comparison meaningful with ANT. For the fixed neural networks, we compare two variants, ones that are fully connected (FC) and ones that are partially connected (PC), with a maximum of nine connections feeding into one neuron.

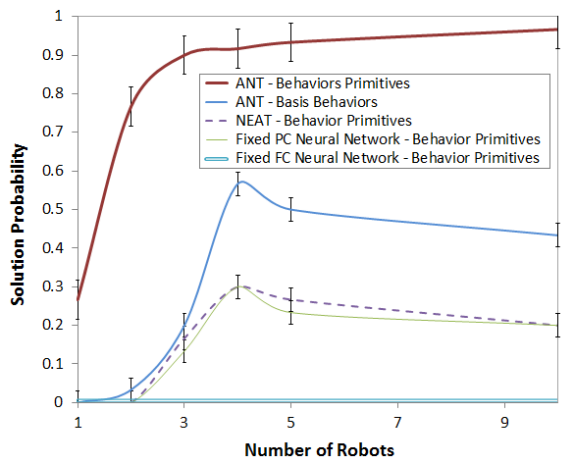

*Figure 7: Probability of obtaining a solution (fitness >0.9) for the resource-collection task using various neural network control approaches. ANT based controllers, particularly using behavior primitives, show the best performance.*

The worst results are obtained with the fully connected neural networks, where fitness performance is lower than all other cases and finds no solutions (fitness > 0.9). Fully connected neural networks are faced with the problem of spatial crosstalk, where spurious signals can drown out signals from important neurons making training difficult or intractable. Similarly, partially connected fixed and variable topology networks also tend to have more `active' synaptic connections present (since all neurons are active) and thus takes longer for each neuron to tune these connections to the sensory inputs. ANT is an improvement as the decision neurons learn to actively mask out spurious neurons producing fitter solutions compared to conventional neural networks.

### 5.2 Scalability Comparison

Having found that ANT can find fitter solutions to the resource collection task than conventional methods, we analyze the fittest ANT solutions (using behaviors primitives) from the simulation runs for scalability in the number of robots while holding the amount of resources constant (Figure 8). The results for ANT using basis behaviors are presented in [10].

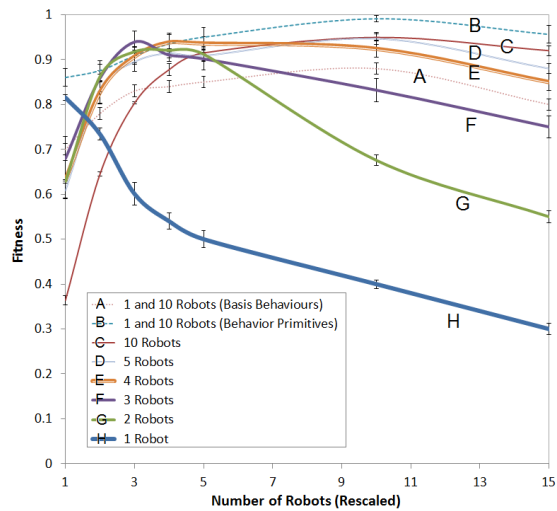

*Figure 8: Scalability performance of evolved robot controllers. The typical results show low performance for fewer robots because it is understaffed for the task at hand. Beyond an optimal density, too many robots result in antagonism, steadily reducing system performance.*

The results for behavior primitives show an overall improvement in fitness compared to those with basis behaviors, though the overall trends remain the same. Taking the controller evolved for a single robot, and running it on a multirobot system shows limited improvement. Using four or more robots results in a gradual decrease in performance due to the increased *antagonism* created. It is found that the scalability of the evolved solution depends on the number of robots used during training. The single-robot controller expectedly lacks the cooperative behavior necessary to function well within a multiagent setting. For example, such controllers fail to develop robot collision avoidance or bucket brigade behaviors.

The robot controllers evolved with two or more robots perform worse when scaled down to a single robot, showing that the solutions are dependent on cooperation among robots. Hybrid solutions that explicitly train on boundary conditions, including one-robot scenario and maximal n-robot scenario, show the best scalability.

### 5.3 Evolution of Bucket Brigade Behavior

Some of the evolved solutions indicate that the individual robots figure out how to dump resources into the designated dumping area, but not all robots deliver resources all the way to the dumping area every time. Instead, the robots learn to pass the resource material from one individual to another during an encounter, forming a 'bucket brigade' (see Figure 6). Figure 9 shows the probability of obtaining a bucket brigade solution for the resource collection task. An increased number of robots increases the probability ANT obtains a bucket brigade solution. This intuitively makes sense because an increased number of robots would require robots to act in a coordinated fashion to assemble a bucket brigade. Too few robots decrease the chance of robot-robot encounters, making it unlikely for the bucket brigade behavior to be evolved.

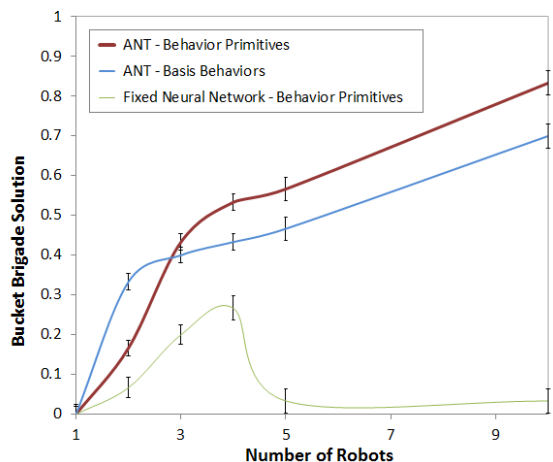

Figure 9: The probability of controllers using a bucket brigade solution versus the number of robots. ANT based controllers learn to utilize the bucket brigade strategy with an increased number of robots because it increases the efficiency of the group in transporting resources to the dumping area.

A bucket brigade approach is beneficial because it increases the efficiency of the group. The robots are effectively designated to a limited area, where they gather resources and transport it to the neighboring robot until the resources are gathered to the dumping area. ANT manages to exploit the bucket brigade strategy while fixed neural networks are found unable for an increase in the number of robots. Incidentally, fixed neural networks show limited scalability for an increased number of robots. Both of these factors go hand in hand. ANT with behavior primitives is more effective than ANT with basis behaviors at using a bucket brigade solution. Behavior primitives provide a larger repertoire of behaviors to select from through trial and error learning.

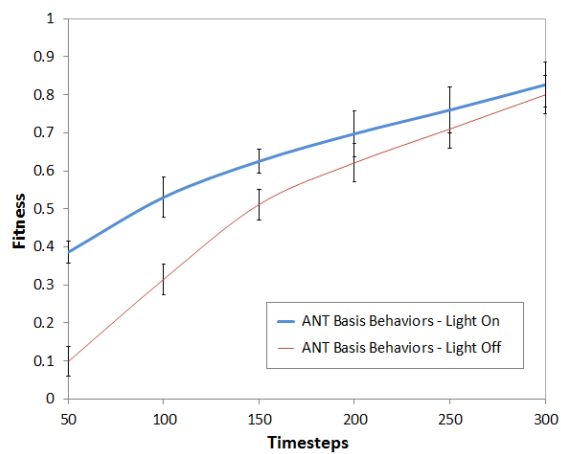

Figure 10: These results show ANT based controllers can exploit phototaxis behavior and evolve to use the light beacon for navigation when time is limited.

### 5.3 Evolving to Exploit Environment Cues

Phototaxis is the process of homing in on a light source. It is an example of a template in the environment. Homing to a light source is widely used by insects and single-cell organisms. Phototaxis requires a means to determine the light gradient and intensity, both of which are made available to each robot. The robot model assumes the light-detection sensor is directional and has a limited field of view. Hence, once a robot faces away from the light source, the light detection sensor is in the ``Not Visible" state. Figure 10 shows the results of ANT with the light beacon turned on and off. The results show that light homing behavior is being used when there is a short time available to complete the task resulting in substantial performance improvement. Homing in on the light beacon helps in navigation by allowing the robots to travel directly to the dumping area.

We also find that templates may not always provide a net benefit to a team of decentralized robots. In one instance, we attempted partitioning the workspace into equal-sized areas demarcated using blue. The blue boundaries formed a grid pattern that divided up the work area into one region, four equal-sized regions, or 16 equal-sized regions (Table 3). By partitioning the space, we would expect each robot to stick to its work-area and minimize interaction with other robots. Under this scenario, we would have robots that traverse along the blue boundaries to carry material to the dumping area and other robots that collect inside boundaries. It turns out such partitioning introduces more complexity to the environment, thus resulting in reduced training performance.

Table 3: Effect of Template Partitions on Fitness

| Scenario | Fitness |
|---|---|
| 1 Template Partition, 4 robots | 71 ± 3 |
| 4 Template Partitions, 4 robots | 59 ± 3 |
| 16 Template Partitions, 4 robots | 49 ± 3 |

## 6 CONCLUSIONS

A neural network architecture called ANT has been evaluated for a multi-robot resource collection task. In comparison, conventional fixed and variable topology neural networks tested were unable to find robust solutions. Because little preprogrammed knowledge is given, an ANT architecture may discover creative solutions that might otherwise be overlooked by a human supervisor. ANT controllers are shown to evolve the ability to exploit templates such as light-beacons and simple cues. Overall, ANT controllers exhibit improved scalability over conventional methods and can reduce the effects of antagonism due to increased robot density. Evolved controllers are found to produce a wide spectrum of group behaviors for varying robot densities and time constraints.